\documentclass[letterpaper, 10 pt, conference]{ieeeconf}  

\IEEEoverridecommandlockouts                              

\usepackage{graphicx} 
\usepackage{tikz}
\def\checkmark{\tikz\fill[scale=0.4](0,.35) -- (.25,0) -- (1,.7) -- (.25,.15) -- cycle;} 
\usepackage{multirow}

\usepackage{xcolor}
\usepackage{graphicx}
\let\labelindent\relax
\usepackage{enumitem}

\usepackage{float}
\usepackage{cite}

\usepackage{algorithm, algorithmicx, algpseudocode}
\usepackage{siunitx}
\usepackage{booktabs}

\usepackage{stfloats} 

\definecolor{lightred}{rgb}{0.9, 0.2, 0.2} 
\definecolor{yellow}{rgb}{0.90, 0.7, 0.1}  

\definecolor{lightblue}{rgb}{0.4, 0.6, 1.0} 
\definecolor{red}{rgb}{1.0, 0.3, 0.3}       
\definecolor{yellow}{rgb}{1.0, 0.85, 0.2}   
\definecolor{purple}{rgb}{0.6, 0.1, 0.6}    

\title{\LARGE \bf
Immersive Social Interaction with VR and LLM-Assisted Humanoids
\author{Niraj Pudasaini$^{1*}$, Geeta Chandra Raju Bethala$^{1*}$, Pranav Doma$^{1*}$, Anthony Tzes$^{1}$, Yi Fang$^{1}$%
\thanks{$^{1}$ All authors are affiliated with New York University Abu Dhabi (NYUAD), UAE. 
        {\tt\small \{np2289, gb2643\}@nyu.edu}}}
}

\begin{document}

\maketitle
\thispagestyle{empty}
\pagestyle{empty}



\section{Introduction}

The growing demand for human-robot interaction in diverse fields such as elder care, social engagement, and search-and-rescue has driven significant advancements in teleoperation systems for humanoid robots. 
The aging global population has led to a rise in social isolation and mobility challenges, especially for homebound older adults \cite{NAP25663}. This isolation significantly affects the mental and physical well-being of aging people, creating a need for innovative solutions that facilitate remote engagement and interaction. Conversational telepresence and teleoperation robots offer a promising tool for these individuals to connect with the external world, participate in social activities, and maintain a sense of independence.  Beyond enhancing social interaction for older adults, teleoperation robotic systems can potentially be used for search and rescue operations in hazardous environments \cite{10876011} where human presence is dangerous. However, existing teleoperated robotic systems often face significant challenges in locomotion control, particularly when navigating complex and unstructured environments \cite{hokayem2006teleoperation, darvish2023teleoperation}. Additionally, traditional teleoperation interfaces for controlling robot locomotion are less user-friendly and cognitively demanding \cite{rea2022usercentered}, requiring operators to manage multiple joints simultaneously. 

This paper introduces a versatile humanoid teleoperation system that leverages voice commands for locomotion and VR hand tracking for manipulation, enabling users to have whole-body control of a humanoid robot for \textit{locomotion}, \textit{manipulation} and \textit{social interaction} tasks. Additionally, Our framework can be used for multi-modal data collection during teleoperation, which can then be utilized to train robots via imitation learning for more complex tasks. Beyond the tasks studied here, the proposed interface can support natural human-humanoid interactions and collaboration scenarios such as cooperative manipulation \cite{11203041}.


\begin{figure}[htb!]
\centering
\includegraphics[width=.99\linewidth]{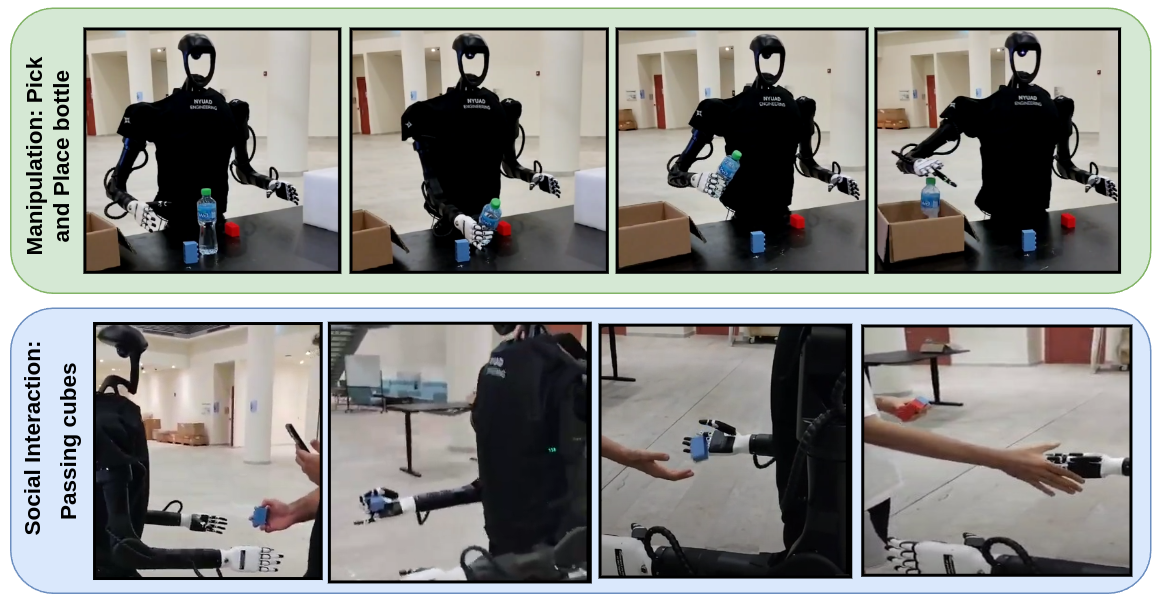}
\caption{Demonstration of tasks involving voice-controlled locomotion, teleoperated manipulation, and social interaction. For the \textbf{manipulation} task, the user teleoperates with the hands to pick the bottle and place it in a box. For the \textbf{social interaction} task, the robot takes the blue cube from a person, walks towards another person, and hands the cube. Finally, the robot performs hand-shaking gesture.}
\vspace{-20pt}
\label{fig:tasks}
\end{figure}


\begin{figure*}[h!]
\centering
\includegraphics[width=\textwidth]{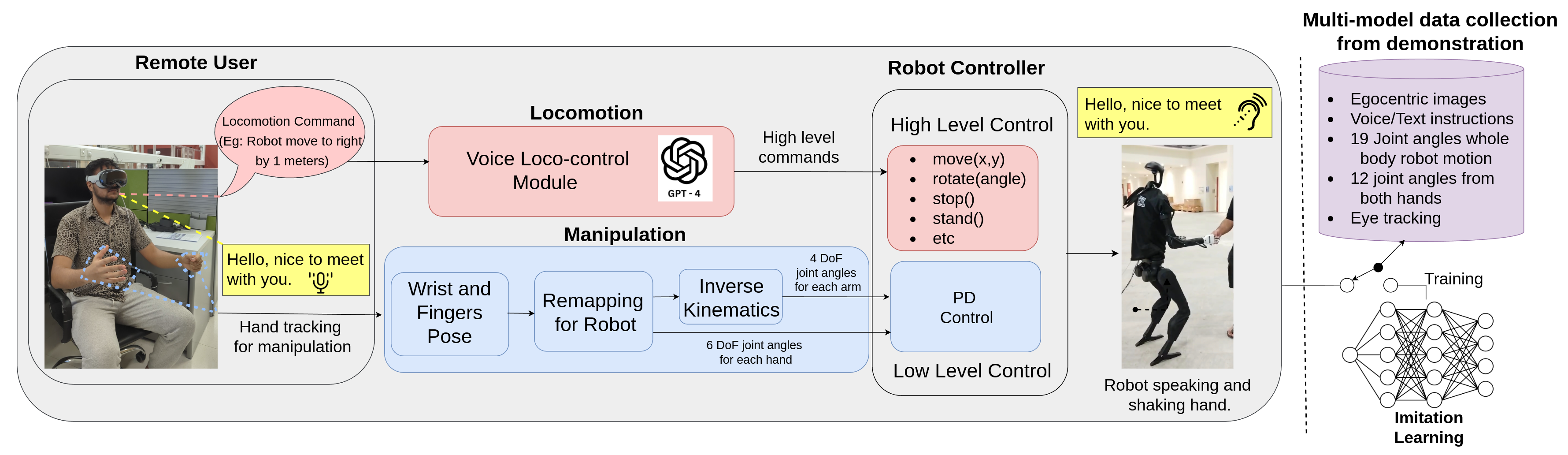}
\caption{Our system features \textcolor{red}{voice-controlled locomotion}, \textcolor{lightblue}{teleoperated manipulation}, \textcolor{yellow}{social interaction}, and multi-modal data collection ability. The user uses Apple Vision Pro to send voice commands for locomotion, social interaction, and hand tracking. The voice loco-control module converts locomotion commands into high-level control commands, while the manipulation module tracks wrist and finger poses to send low-level commands via PD control. Multi-modal data, \textit{i.e.}, ego-centric images, voice/text commands, 19 body and 12 hand joint angles, and eye movements, is recorded for imitation learning.}
\vspace{-20pt}

\label{fig:met}
\end{figure*}

\section{Methodology}
Our method consists of: 1). voice-controlled locomotion, 2). teleoperation manipulation, and 3). social interaction for a humanoid robot, facilitating real-time whole-body control and allowing multi-modal data collection for task learning (see Fig \ref{fig:met}). 


\subsection{Voice Commands for High-level Locomotion}



Based on the ego-centric images streamed with a resolution of 640 $\times$ 480 on an Apple Vision Pro, the user can send the locomotion command through voice. The voice loco-control module translates the locomotion commands into high-level control commands, such as \textit{move(x, y), rotate(angle), stop()}, and \textit{stand()}. Then, the relevant high-level functions are called for robot navigation. The robot's bi-pedal locomotion is based on a pre-trained deep reinforcement learning model to produce robust locomotion policy, as developed in prior works \cite{expressive2023, kumar2021rma}. Note that during implementation, we faced challenges with the robot's bi-pedal morphology, which does not inherently stabilize itself.


For the voice loco-control module, we use Deepgram \cite{deepgram} for real-time speech-to-text transcription, GPT-4's \cite{openai2023gpt4} reasoning to parse locomotion commands from text into high-level control commands, and Silero \cite{SileroModels} for text-to-speech synthesis with the LivKit \cite{livkit} agent framework. During implementation GPT-4 sometimes misinterprets human voice commands. To mitigate this, we implemented a verification step where, if GPT-4 finds the instruction uncertain, the system asks the user to confirm or clarify the command before execution.

If the user wants the scene description, it can be generated on-demand using GPT-4's vision capabilities, but remain disabled by default to conserve computational resources and optimize inference speed. 


\subsection{Teleoperation for Manipulation}
We adopt Apple Vision Pro \cite{apple2023visionpro} and VisionPro Teleop \cite{park2024avp} to stream the human operator's wrist and finger poses in SE(3) to the server installed on the robot. The streamed data is then re-targeted and re-mapped to the robot's arms and fingers that follow the operator's movements. 

We use Inspire Robotics \cite{inspire2024dexteroushands} dexterous hands as end effectors. For teleoperation manipulation, we control 4 DoF for each arm and 6 DoF for fingers of each hand. Human wrist poses are transformed into the robot's coordinate frame, and we use inverse kinematics from Pinocchio \cite{pinocchio2024clik} \cite{carpentier2021pinocchio} to compute the joint angles and then pass through the PD controller to make robot's arm mimics the operator's arm. 


\subsection{Social Interaction}

We implemented bidirectional audio  transmission using the ROS 1\cite{quigley2009ros}. ROS nodes on both the operator and robot enable real-time audio streaming, allowing the operator to hear the robot's environment and communicate with people around the robot via the robot's speakers. This setup enhances situational awareness and supports effective telepresence \cite{rea2022usercentered} for social interaction. We found that only egocentric view is not suitable for navigation; however, for future work, we plan to integrate an additional waist-mounted camera to enhance the operator's situational awareness.






\section{Experiment Setup}

\begin{figure}[htbp]
    \centering
    \includegraphics[width=0.45\textwidth]{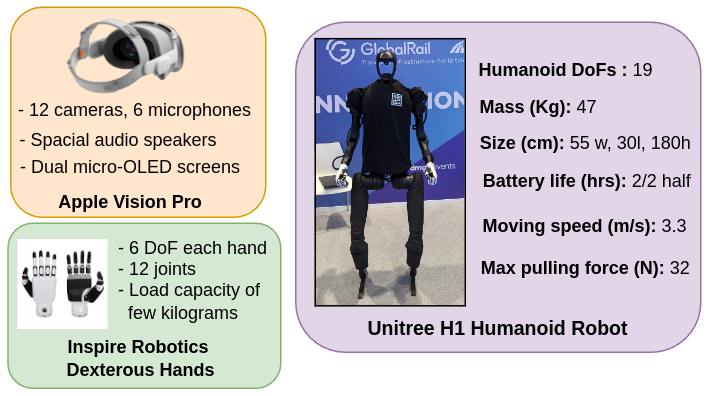}
    \caption{Apple Vision Pro, Inspire Robotics Dexterous Hands, and Unitree H1 Humanoid Robot used in the experimental setup.}
    \label{fig:experiment_setup}
    \vspace{-10pt}
\end{figure}

The experiment is setup with to evaluate our humanoid teleoperation system's capabilities in locomotion, manipulation, and social interaction (see Figure\ref{fig:tasks}).  The hardware details are shown in Figure \ref{fig:experiment_setup}

\begin{itemize}
    \item \textbf{Manipulation Task (Pick and Place Bottle):} This task involves teleoperating the robot to pick up a bottle placed randomly on a table and place it into a box. Successful completion demonstrates the robot's precise manipulation and locomotion control as it navigates to the bottle, grasps it securely, and transports it to the target location.
    
    \item \textbf{Social Interaction Task (Passing Cube):} The robot needs to verbally request a cube from one person through verbal communication. It then walks to another person and hands over the cube, ensuring transfers occur based on clear instructions. This task showcases the robot's coordination of locomotion, manipulation, and social cues for effective social interaction.
\end{itemize}


\section{Results and Discussions}


\begin{table}[!ht]
\caption{Performance Comparison of New \textit{vs.} Expert Users}
\label{table_performance}
\centering
\vspace{-0.2cm}
\resizebox{0.8\columnwidth}{!}{%
\begin{tabular}{c c c c c}
\toprule
\multirow{2}{*}{\textbf{Task}} & \multicolumn{2}{c}{\textbf{Success Rate}} & \multicolumn{2}{c}{\textbf{Time (s)}} \\
\cmidrule(r){2-3} \cmidrule(r){4-5}
 & \textbf{Novice} & \textbf{Expert} & \textbf{Novice} & \textbf{Expert} \\
\midrule
Object Pick & 0.8 & 0.90 & 52 & 22 \\
Social Interaction & 0.7 & 0.8 & 326 & 158 \\
\bottomrule
\end{tabular}
}
\vspace{-0.2cm}
\end{table}

Table \ref{table_methods} highlights that our method uniquely supports voice-controlled locomotion, manipulation and social interaction, unlike existing approaches such as Human Plus\cite{fu2024humanplushumanoidshadowingimitation} and Human to Humanoid\cite{he2024learninghumantohumanoidrealtimewholebody}. This makes our system more accessible and less physically demanding, particularly beneficial for homebound older adults and applications in confined or hazardous environments. Additionally, the multimodal data collected during teleoperation can support downstream autonomy beyond imitation learning: aligned voice/text instructions, egocentric observations, and executed robot actions can provide priors for embodied action reasoning in humanoid loco-manipulation \cite{wen2025humanoid}, while egocentric RGB streams can support high-fidelity 3D scene reconstruction for perception and mapping using Gaussian Splatting techniques \cite{deng2025hierarchical}.

\begin{table}[!htb]
\caption{Comparison of Humanoid Teleoperation Methods}
\label{table_methods}
\centering
\scriptsize
\resizebox{\columnwidth}{!}{
\begin{tabular}{c c c c}
\toprule
\textbf{Method} & \textbf{Locomotion} & \textbf{Manipulation} & \textbf{Social Interaction} \\
\midrule
Open Television \cite{cheng2024opentelevisionteleoperationimmersiveactive} &  & $\checkmark$ &  \\
Human Plus \cite{fu2024humanplushumanoidshadowingimitation} & $\checkmark$ & $\checkmark$ &  \\
Human to Humanoid \cite{he2024learninghumantohumanoidrealtimewholebody} & $\checkmark$ & $\checkmark$ &  \\
\hline
\textbf{Ours} & $\checkmark$ & $\checkmark$ & $\checkmark$ \\
\bottomrule
\end{tabular}
}
\end{table}

\section{Conclusion}
This paper introduces a comprehensive humanoid teleoperation system integrating voice-controlled locomotion, VR-based manipulation, and enhanced social interaction capabilities. The system addresses the limitations of existing approaches by enabling intuitive, whole-body control and social engagement, crucial for elderly care and hazardous environments. The capacity of multi-modal data collection lays the groundwork for future advancements in imitation learning, improving the robot's autonomy and task proficiency. Future work includes refining the social interaction module and enhancing multi-modal data processing for more sophisticated behavior learning.

\bibliographystyle{IEEEtran}
\bibliography{biblio}

\end{document}